\documentclass[letterpaper]{article} 
\usepackage{aaai24}  
\usepackage{times}  
\usepackage{helvet}  
\usepackage{courier}  
\usepackage[hyphens]{url}  
\usepackage{graphicx} 
\usepackage{natbib}  
\usepackage{caption} 
\usepackage{utfsym}
\usepackage[linesnumbered,ruled,vlined]{algorithm2e}
\SetCommentSty{small}
\usepackage{amsmath}
\usepackage{amssymb}
\usepackage{booktabs}
\usepackage{multirow}
\usepackage{xcolor}
\usepackage{enumitem}

\title{FedMKGC: Privacy-Preserving Federated Multilingual Knowledge Graph Completion}
\author {
    Wei Tang\textsuperscript{\rm 1},
    Zhiqian Wu\textsuperscript{\rm 1},
    Yixin Cao\textsuperscript{\rm 2},
    Yong Liao\textsuperscript{\rm 1},
    Pengyuan Zhou\textsuperscript{\rm 1},
}
\affiliations {
    \textsuperscript{\rm 1}University of Science and Technology of China, Hefei, China \\
    \textsuperscript{\rm 2}Singapore Management University, Singapore\\
    weitangcs@gmail.com,
    wuzhiqian@mail.ustc.edu.cn,\\
    caoyixin2011@gmail.com,
    yliao@ustc.edu.cn,
    zpymyyn@gmail.com
}

\usepackage{bibentry}

\nocopyright

\begin{document}

\maketitle

\begin{abstract}
   Knowledge graph completion (KGC) aims to predict missing facts in knowledge graphs (KGs), which is crucial as modern KGs remain largely incomplete. While training KGC models on multiple aligned KGs can improve performance, previous methods that rely on transferring raw data among KGs raise privacy concerns. To address this challenge, we propose a new federated learning framework that implicitly aggregates knowledge from multiple KGs without demanding raw data exchange and entity alignment. We treat each KG as a client that trains a local language model through text-based knowledge representation learning. A central server then aggregates the model weights from clients. As natural language provides a universal representation, the same knowledge thus has similar semantic representations across KGs. As such, the aggregated language model can leverage complementary knowledge from multilingual KGs without demanding raw user data sharing. Extensive experiments on a benchmark dataset demonstrate that our method substantially improves KGC on multilingual KGs, achieving comparable performance to state-of-the-art alignment-based models without requiring any labeled alignments or raw user data sharing. Our codes will be publicly available.    
    \end{abstract}
    
    \section{Introduction}    
    
    Knowledge Graphs (KGs), which contain extensive structural facts in the form of triples \texttt{(head entity, relation, tail entity)}, are essential for various knowledge-driven applications, such as question answering \cite{yasunaga-etal-2021-qa} and recommendation \cite{10.1145/3292500.3330989}. More recently, KGs are also used to supplement the knowledge \cite{sun2021ernie} and enhance the reasoning ability \cite{hu-etal-2022-empowering} of large language models (LLMs).
    However, almost all KGs, even those at the scale of billions of triples, are far from complete. Therefore automatically predicting missing facts in KGs, also known as knowledge graph completion (KGC), is crucial for KGs to be used in real-world applications.

    In the multilingual setting, the incompleteness of KGs is even more severe especially for low-resource language. Due to the sparseness, the monolingual KGC methods that learn each KG separately perform badly in low-resource KGs \cite{tong-etal-2022-joint}. 
    However, it is notable that the knowledge of different language-specific KGs can be complementary. Hence, transferring knowledge from rich-resource KGs to low-resource KGs can benefit multilingual KGC.
    Recent studies \cite{chen2020multilingual, singh2021multilingual, huang2022multilingual, tong-etal-2022-joint} highlight significant advancements in multilingual KGC (MKGC) by explicit knowledge aggregation. They create connections between multiple KGs using aligned entities. Through these interconnections, raw data and knowledge representations are shared and transferred between KGs.

    \begin{figure}[t]
        \centering
        \includegraphics[width=0.46\textwidth]{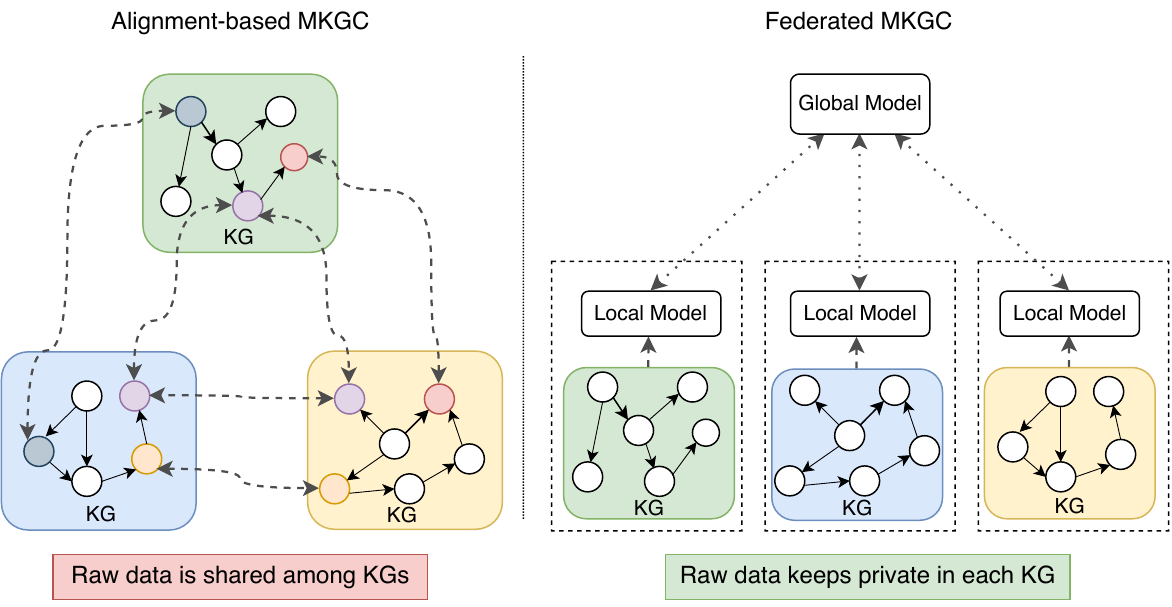} 
        \caption{Previous MKGC methods rely on alignment, which requires raw data sharing among KGs. Our proposed federated MKGC keeps data private in each KG.}
        \label{fig:intro}
    \end{figure}

    While present MKGC methods exhibit promise, a common underlying assumption persists: the KGs can freely exchange raw data. This assumption is problematic in real-world applications. In practice, KGs often form the foundational assets of organizations, encapsulating sensitive and proprietary data. This is particularly evident in domains such as medical and financial where KGs may hold significant personal and private information. Given the rising concerns over data privacy, underscored by regulations like the European Union General Data Protection Regulation, the direct sharing of raw KG data is typically constrained to ensure privacy protection. 
    Additionally, achieving entity alignment in practical scenarios is challenging. Relying predominantly on human annotation, this alignment is not only labor-intensive but also time-consuming. And the alignment process is also exposed to a huge risk of privacy leakage, since it requires KGs to share almost all entities with others.

    To address the aforementioned challenges, we resort to Federated Learning \cite{mcmahan2017communication}, ensuring that raw data remains unshared among KGs. Under the federated setting, these KGs collaboratively train a shared model with a focus on privacy, eliminating the need for raw data exchanges and alignment annotations.
    For efficient MKGC in this framework, we move away from embedding-based methods, which aim to derive optimal embeddings for known entities and relations. Such methods inherently result in a transductive manner, wherein knowledge transmission depends on data exchange, which conflicts with the federated paradigm. In contrast, we advocate for inductive knowledge graph embedding learning, such as text-based KGC methods. These methods intrinsically embed knowledge into the model's parameters, allowing knowledge from multiple KGs to be aggregated implicitly by merging the parameters of various models, negating the need for raw data exchange or alignment annotation.
    
    Motivated by these observations, we propose FedMKGC, a new \textbf{Fed}erated learning framework for \textbf{M}ultilingual \textbf{KGC}. This approach implicitly aggregates knowledge across multiple KGs by training a global language model. Negating data transmission among KGs, FedMKGC directly bridges knowledge from local KGs to the global language model, thereby enabling complementary knowledge utilization circumventing privacy-concerned data sharing and time-consuming alignment annotations, achieving superior KGC performance in multilingual scenarios.
    To be specific, FedMKGC lets each client possesses their own KG. These clients train local language models using text-based knowledge representation learning. Intrinsic to these local language models are weights that encapsulate the respective knowledge. These weights are subsequently uploaded to a central server, which then assimilates the complementary knowledge extracted from disparate clients via weight aggregation.
    For text-based knowledge representation learning, knowledge triples are first converted into natural language texts. The main goal of this representation learning is to predict the correct tail entity (or head entity) when presented with an incomplete triple that includes a head entity (or tail entity) and the associated relation. To facilitate this, we introduce a contrastive learning approach to demand language model generate close representations for incomplete triple and the corresponding correct entity pairs. Notably, this method holds promise for inductive KGC as it integrates all knowledge into the language model's parameters, thereby enabling it to assimilate information from other KGs.
    We conduct comprehensive experiments on a popular public multilingual benchmark. The results show that FedMKGC substantially improves KGC on multilingual KGs. When compared to the state-of-the-art models that rely on aligned entities, FedMKGC achieves comparable performance with its elimination of annotation requirements and its inherent privacy safeguards.
    We summarize our contributions as follows:    
    \begin{itemize}
      \item We propose an innovative federated learning framework, FedMKGC, for aggregating knowledge from multilingual KGs, bypassing aligned entities and eliminating the need for time-consuming annotations and privacy concerns.
      \item  We introduce a text-based knowledge representation approach using natural language, combined with a contrastive learning method, to embed knowledge into the parameters of the language model.
      \item  Extensive experiments conducted on a popular multilingual benchmark demonstrate the significant improvements brought by FedMKGC with further investigation on the framework design.
    \end{itemize}
    
    \section{Related Work}
    
    \subsection{Monolingual Knowledge Graph Completion}
    
    In prior stages, a series of embedding-based models like TransE~\cite{bordes2013translating}, TransH~\cite{wang2014knowledge}, TransR~\cite{lin2015learning}, and RotatE~\cite{sun2019rotate} have been proposed. These models translate entities and relations into high-dimensional vector spaces, employing score functions for triples $(h,r,t)$ to facilitate KGC.
    Besides, there are also various methods such as RESCAL~\cite{nickel2011three},  DistMult\cite{yang2014embedding}, HolE \cite{10.5555/3016100.3016172}, and TuckER\cite{balazevic-etal-2019-tucker} introduce novel methods based on tensor factorization.

    In addition to embedding-based KGC methods, researchers also dive deeply into text-based methodologies for the potential of inductive knowledge representation, which provides the capability to model entities not encountered during the training phase. Besides utilizing the structural information, StAR~\cite{wang2021structure}, GenKGC~\cite{xie2022discrimination}, KRACL~\cite{tan2023kracl} introduce additional textual representations to improve the performance of KGC. In a parallel vein, KG-BERT~\cite{yao2019kg}, PKGC~\cite{lv2022pre}, SimKGC~\cite{wang-etal-2022-simkgc} employ fine-tuned pre-trained models to obtain semantic representations of entities and relations for KGC.

    \subsection{Multilingual Knowledge Graph Completion}

    Recently, researchers have extended the KGC methods to bridge knowledge from multiple KGs with the purpose to foster the advancement of multilingual KGC.
    MTransE~\cite{chen2016multilingual} encodes multilingual knowledge graph triples in high-dimensional vector spaces, aligning embeddings of corresponding counterparts. KEnS~\cite{chen2020multilingual} facilitates cross-lingual knowledge transfer across distinct language-specific knowledge graphs. ALIGNKGC~\cite{singh2021multilingual} leverages alignment seeds among entity pairs in multilingual KGs, jointly tackling KGC, entity alignment, and relation alignment. JMAC~\cite{tong-etal-2022-joint} employs Graph Neural Networks (GNNs) to reinforce structural consistency in multilingual KGs. SS-AGA~\cite{huang2022multilingual} and CKGC-CKD~\cite{zhang2023collective} combine KGs into a unified graph, treating alignment as a novel edge type. 
    While extant methods in the domain of MKGC have undoubtedly demonstrated promising performance, it is important to note that these approaches heavily depend on explicit entity alignment. This dependency necessitates substantial manual annotation efforts, and also engenders pertinent privacy concerns. Consequently, these constraints considerably limit the practical usability of these methods. 
    
    \subsection{Federated Learning}
    
    Federated learning \cite{mcmahan2017communication} is proposed to ensure privacy when implementing machine learning methods involving distributed clients. Clients independently train local models and share aggregated parameters with a central server. In the pursuit of safeguarding privacy, an emerging avenue of research has delved into the domain of KGC within the framework of federated learning. FedE~\cite{FedE} embeds KGs locally, aggregates entity embeddings on a server, and dispatches corresponding entity embeddings to clients to protect KG facts' privacy. FedEC~\cite{CHEN2022109459} introduces contrastive learning to handle data heterogeneity using FedE. FedLU~\cite{zhu2023heterogeneous} employs unlearning and knowledge distillation to mitigate data heterogeneity and knowledge forgetting. Furthermore, FKGE~\cite{Peng-2021-DPFKGE} introduces adversarial generation between pairs of KGs for embedding aligned entities and relations in a federated setting. However, the present KGC methods within the paradigm of federated learning are marked by a notable reliance on aligned entities, this dependency curtails the broader practical utility of these methods. Contrastively, our proposed approach stands out by obviating the need for labor-intensive manual annotations or the exchange of raw data, which is vital for MKGC in real-world applications.

    \section{Methodology}

    In this section, we introduce FedMKGC, a novel federated learning framework designed for multilingual knowledge graph completion. Within FedMKGC, each client possesses its own KG and independently trains a language model by text-based knowledge learning. Through it, These language models embed the knowledge in the individual KGs into their model weights. A central server subsequently aggregates the model weights from these local language models, facilitating the implicit integration of knowledge across these multilingual KGs without the necessity for direct raw data sharing.

    Specifically, FedMKGC employs a federated learning architecture with each KG acting as a client. The local language model in each client is trained to embed the knowledge triples in its KG. The model weights are then aggregated at a central server to consolidate the knowledge in an integrated way, while keeping the raw KG data decentralized.
    For knowledge representation learning, FedMKGC commences by converting the knowledge triples from each KG into natural language texts. Subsequently, a language model is employed to produce distributed representations of these triples. This is executed via a contrastive learning approach, which requires the language model to generate close representations for element-pairs that constitute correct triples. 
    
    \subsection{Problem Definition}
     A KG is denoted as $G = (\mathcal{E, R, T})$, where $\mathcal{E}$, $\mathcal{R}$ and $\mathcal{T}$ denote the set of entities, relations, and fact triples, respectively. Each fact triple is represented as $(h, r, t)$, where $h, t \in \mathcal{E}$ and $r \in \mathcal{R}$.
     The goal of KGC is to predict the tail entity $t$ given the head entity $h$ and relation $r$ as a query $(h, r, ?)$, or predict the head entity $h$ given the tail entity $t$ and relation $r$ as a query $(?, r, t)$.
     For instance, given the query (``Sherlock Holmes'', ``lives in'', ?), the KGC model should predict the tail entity ``London''.

     MKGC aims to propagate knowledge across multilingual KGs and utilize complementary information to enhance each KG's completion performance. We denote the KGs in M languages as $G_1, G_2, \cdots, G_M$, and the corresponding entity set as $\mathcal{E}_1, \mathcal{E}_2, \cdots, \mathcal{E}_M$.
     Previous methods depend on explicitly aligned entities across KGs for knowledge transfer, assuming an abundance of such aligned entities is already available. However, we contend that obtaining such aligned entities in practice is far from trivial. It is a process that is both time-consuming and labor-intensive. Additionally, labeling aligned entities necessitates the exchange of raw data between KGs, which could raise privacy issues when the KGs are owned by different organizations.
     Therefore, our primary focus is on scenarios where no aligned entities are provided. In these cases, we must rely solely on the implicit knowledge within KGs to facilitate knowledge transfer between them.

    \subsection{Federated Learning for Multilingual Knowledge Graph Completion}

    We introduce a novel federated learning framework tailored for privacy-preserving multilingual knowledge graph completion. This framework hinges on the principle of implicit knowledge aggregation. Specifically, each client operates its own KG, and sharing raw data between clients is not allowed for privacy safeguarding. Instead, each client locally trains a language model on its KG through text-based knowledge representation learning—further elaborated in the subsequent section. This process embeds the knowledge directly into the local language model's weights. We employ a weight aggregation strategy to amalgamate the embedded knowledge from different clients without the need for direct raw data exchange. This also eliminates the cumbersome and laborious process of alignment annotations.

    Structurally, our framework parallels general federated learning architectures and is segmented into four steps:
    \begin{enumerate}[label=(\roman*), align=left]
    \item The server initializes the weights of the global language model.
    \item During each round, the server selects $m \leq M$ clients for local language model training.
    \item The selected clients update the latest weights received from the server into their local language models. Post this integration, they then continue to train with the local KG and relay the updated weights back to the server.
    \item The server then harnesses a weight aggregation mechanism to refine the global language model.
    \end{enumerate}
    Steps (ii) through (iv) recur until a predetermined number of rounds is met. Details of this procedure can be seen in Algorithm~\ref{alg:fl}.
    
    Knowledge consolidation is accomplished when the server aggregates the weights from local language models. Formally, the widely-used weighted average aggregation mechanism is defined as follows:
     \begin{equation}
     \label{Fedavg}
         w_{t+1} =  \sum\limits_{i \in S_{t}} \frac{n_{i}}{|S_{t}|} w_{t}^{i},
     \end{equation}
    where $w_{t+1}$ denotes the aggregated weights that local clients would receive from the server in round $t+1$. $S_{t}$ is a set of selected clients in round $t$. $w_{t}^{i}$ denotes the local weights in the $i$-th selected client in round $t$. $\frac{n_{i}}{|S_{t}|}$ stands as the proportion of $ w_{t}^{i}$ with $\sum\limits_{i \in S_{t}} n_{i}= |S_{t}|$. In our implementation, we regard each KG equal and replace $\frac{n_{i}}{|S_{t}|}$ with $\frac{1}{|S_{t}|}$. 
    These aggregated weights, which implicitly integrate knowledge from the local clients, leverage the complementary information amongst the multilingual KGs enabling augmented KGC capabilities of the global language model.
    
    \begin{algorithm}[h]
    \caption{\text{Federated Learning for MKGC}}
    \label{alg:fl}
    \SetAlgoLined
    \KwIn{KGs in $M$ languages as $G_1, G_2, \ldots,G_M$ on $M$ clients respectively, number of rounds $T$, number of selected clients set $S_{t}$ in round $t$}
    Initialize $w_0$ on server\;
    \For{$t = 1$ \KwTo $T$}{
        \For{each client $i$}{
        Receive aggregated weights $w_{t}$ from server\;
        Train local language model $\text{LM}_i$ on client $i$\;
        Send updated local weights $w_{t}^{i}$ back to server\;
          }
        \tcc{Aggregate local weights on server}
      $w_{t+1} =  \sum\limits_{i \in S_{t}} \frac{n_{i}}{|S_{t}|} w_{t}^{i}$\;
    }
    \end{algorithm}
    
    \subsection{Text-based Knowledge Representation Learning} 
    We employ a text-based knowledge representation learning method to train the language model, enabling it to incorporate knowledge from multiple KGs into its parameters.
    Initially, the knowledge triples are transformed into natural language representations. After that, we adopt a language model to encode the embedding of knowledge triples, which are trained to find the correct entities of the incomplete triples in a contrastive learning manner. The LM is then used to predict the missing facts to complete the KG.

    \subsubsection{Knowledge Embedding} Following recent advancements in text-based knowledge graph completion methods \cite{wang-etal-2022-simkgc}, we train the language model using contrastive learning, where each knowledge triple is divided into two components: the relation-aware head/tail entity and the target tail/head entity. Specifically, given a knowledge triple $(h, r, t)$, we first encode a relation-aware embedding $\mathbf{e}_{hr}$ for the head entity $h$. The language model $\text{LM}$ encodes $\mathbf{e}_{hr}$ by taking as input a concatenated representation comprising the corresponding relation $r$ and entity $h$, where the textual description of entity $h$ (e.g., the entity name) serves directly as its representation. 
    
    On the other hand, rather than directly using the schema as the relation's representation, we draw inspiration from prefix-tuning \cite{li-liang-2021-prefix} and treat a set of trainable parameters $p_r$ as the parameterized relation representation. We posit that relations characterize the rich ontology of the knowledge graph, which is far more complex than the entities. Using trainable parameters provides greater flexibility than fixed schema descriptions in multilingual knowledge graph scenarios. Since the input sequences are multiple tokens, we use mean pooling followed by $L_2$ normalization to get the final relation-aware embedding of the head entity $\mathbf{e}_{hr}$

    Besides, we add inverse relation $r'$ for each relation $r$, so that the relation-aware embedding of the tail entity $\mathbf{e}_{tr}$ is encoded in the same way as $\mathbf{e}_{hr}$, where the representations of relation and entity are replaced with the inverse relation $p_{r'}$ and textual description of tail entity $t$, respectively.

    After that, we encode the embedding of the target entity, e.g., tail entity $\mathbf{e}_t$ for $\mathbf{e}_{hr}$ and head entity $\mathbf{e}_h$ for $\mathbf{e}_{tr}$, by directly inputting its textual description into the same LM. The embedding is also obtained through mean pooling followed by $L_2$ normalization.

    \subsubsection{Constrative Learning} We adopt contrastive learning to train the LM to generate similar embedding of the relation-aware entity and the corresponding target entity. To be specific, we regard the relation-aware entity embedding as queries and the corresponding target entities as positive samples. And following SimKGC, we conduct in-batch negative sampling to obtain the negative samples. The loss function is defined as follows:
    \begin{equation}
    \label{eq:loss}
        \begin{aligned}
            &\mathcal{L} = \\
            & -\log{\frac{\text{exp}((\mathbf{e}_{x} \cdot \mathbf{e}_{y} - \gamma)/ \tau)}{\text{exp}((\mathbf{e}_{x} \cdot \mathbf{e}_{y} - \gamma)/ \tau)+\sum\limits_{y' \in \mathcal{E'}} \text{exp}(\mathbf{e}_{x} \cdot \mathbf{e}_{y'} / \tau)}}, \\
            & (x, y) \in \{(hr, t), (tr, h)\},  
        \end{aligned}
    \end{equation}    
    where $\mathcal{E'}$ are entities in the same batch with the positive sample, $\tau$ is a temperature parameter, $\gamma>0$ is an additive margin, which motivates the models to elevate the scoring assigned to the correct triple $(h, r, t)$. When process the tail entity prediction $(h,r,?)$, $e_x$ is equivalent to $e_{hr}$, $e_y$ is equivalent to $e_t$, otherwise, in head entity prediction $(?,r,t)$, $e_x$ is equivalent to $e_{tr}$, $e_y$ is equivalent to $e_h$.

    \subsubsection{Prediction} To predict the missing entity in knowledge triple, raising the tail entity prediction $(h, r, ?)$ as the example. We compute the cosine similarity between the $e_{hr}$ and all entities in the entity set of the incomplete client KG $G_i$, and predict the one with the highest score. Following \citeauthor{wang-etal-2022-simkgc}, we also conduct graph-based re-ranking to exhibit the spatial locality. The final score of a candidate entity is added with an additional award if it belongs to the $k$-hop neighbor of the head entity in $G_i$:
    \begin{equation}
        \arg \max_{t_i} \cos(\mathbf{e}_{hr}, \mathbf{e}_{t_i}) + \alpha \mathbb{I}(t_i \in \mathcal{E}_i^k(h)), t_i \in \mathcal{E}_i,
        \label{eq:pred}
    \end{equation}
    where $\alpha$ is a hyperparameter, $\mathbb{I}$ is an indicator function, $\mathbf{E}_i^k(h)$ means the entity set of the $k$-hop neighbor of entity $h$.

    \section{Experiments}

    \begin{table}[h]
        \setlength\tabcolsep{3.2pt}
        \caption{Statistics of the datasets. ``\#Aligned Links'' is the number of aligned entity pairs where one of the entities belongs to that KG.}
        \label{tab:dataset}
        \centering
        \begin{tabular}{l | c c c c}
            \toprule
            Dataset & \#Entities & \#Relations & \#Triples & \#Aligned Links \\
            \midrule
            EN & 13996 & 831 & 80167 & 16916 \\
            FR & 13176 & 178 & 49015 & 16877 \\
            ES & 12382 & 144 & 54066 & 16347 \\
            JA & 11805 & 128 & 28774 & 16263 \\
            EL & 5231  & 111 & 13839 & 9042 \\
            \bottomrule
        \end{tabular}
    \end{table}
    
    \begin{table*}[h]
        \caption{Main results. All metrics are reported in percentage (\%). The highest scores are in bold. ``Private'' denotes whether the model can protect data privacy.} 
        \label{tab:main_results}
        \centering
        \setlength\tabcolsep{2.0pt}
        \begin{tabular}{l | c |  ccc | ccc | ccc | ccc | ccc}
            \toprule
            \multirow{2}*{Model} & \multirow{2}*{Private}  & \multicolumn{3}{c}{Greek} & \multicolumn{3}{c}{Japanese} & \multicolumn{3}{c}{French} & \multicolumn{3}{c}{Spanish} &  \multicolumn{3}{c}{English} \\
            \cline{3-5} \cline{6-8} \cline{9-11} \cline{12-14} \cline{15-17}
            & & H@1 & H@10 & MRR & H@1 & H@10 & MRR & H@1 & H@10 & MRR & H@1 & H@10 & MRR & H@1 & H@10 & MRR \\
            \midrule
            KenS &  \usym{2717} &26.4 &66.1 & -    &32.9 &64.8 & -    &22.3 &60.6 &  -   &25.2 &62.6 &  -   &14.4 &39.6 &  - \\
            SS-AGA &  \usym{2717} &30.8 &58.6 &35.3  &34.6 &\textbf{66.9} &42.9  &25.5 &61.9 &36.6  &27.1 &\textbf{65.5} &38.4  &16.3 &41.3 &23.1 \\
            
            \midrule 
            TransE & \usym{2713} &13.1 & 43.7 &24.3  &21.1 &48.5 &25.3  &13.5 & 45.0 &24.4  &17.5 &48.8 &27.6 &7.3 &29.3 &16.9 \\
            RotatE & \usym{2713} &14.5 &36.2 &26.2 &26.4 &60.2 &39.8 &21.2 &53.9 &33.8 &23.2 &55.5 &35.1 &12.3 &30.4 &20.7 \\
            KG-BERT & \usym{2713} &17.3 &40.1 &27.3  &26.9 &59.8 &38.7  &21.9 &54.1 &34.0  &23.5 &55.9 &35.4  &12.9 &31.9 &21.0 \\
            SimKGC & \usym{2713}  & 22.8 & 46.5 & 26.3  & 21.9 & 39.1 & 28.0  & 20.2 & 40.2 & 27.1  & 17.0 & 36.0 & 23.6  &18.4 & 42.0 & 26.3 \\
            \midrule
            \textbf{FedMKGC} & \usym{2713}  & \textbf{40.1}	& \textbf{73.7}	&	\textbf{52.2}	&	\textbf{35.4}	&	64.6	&	\textbf{45.4}	&	\textbf{34.6}	&	\textbf{64.7}	&	\textbf{45.3}	&	\textbf{34.8}	&	64.0	&	\textbf{45.2}	&	\textbf{30.0}	&	\textbf{61.2}	&	\textbf{40.8} \\
            \bottomrule
        \end{tabular}
    \end{table*}

    \subsection{Dataset}

    We conduct experiments with the DBP-5L \cite{chen2020multilingual} dataset, a multilingual benchmark containing 1392 relations, 56590 entities, and 225831 triples constitute KGs in five different languages: English (EN), French (FR), Greek (EL), Spanish (ES), and Japanese (JA). The statistics of the dataset are shown in Table \ref{tab:dataset}.     

    DBP-5L contains pre-aligned entity pairs among KGs, and on average, about 40\% of the entities in each KG can be aligned to entities in other KGs. Previous works basically rely on these aligned entities to connect multiple KGs, and knowledge is transferred through these connections.
    However, we argue that obtaining such aligned entities is non-trivial in real-world applications. It is a process that is both time-consuming and labor-intensive. Additionally, labeling aligned entities necessitates the exchange of raw data between KGs, which could raise privacy issues when the KGs are owned by different organizations.
    Hence, our primary focus is on scenarios where no aligned entities are provided. In these cases, we must rely solely on implicit knowledge transmission.

    \subsection{Evaluation Protocol}
    In the testing phase, we rank candidate tail entities $\tilde{t}$ in the Knowledge Graph (KG) based on plausibility score $f(h, r, \tilde{t})$ for a given corrupted triple $(h, r, ?)$. We report Mean Reciprocal Rank (MRR), and the proportion of correct predictions in the top-10 (Hits@10) and top-1 (Hits@1) predictions. Similar to the previous works, we also apply the filtered setting, where the candidate entities that are already seen in the training set are removed from the ranking list.

    \subsection{Implementation Details}
    Our experiments are conducted on one NVIDIA V100 32GB GPU. We implement our model with PyTorch and use Adam \cite{DBLP:journals/corr/KingmaB14} as the optimizer. The learning rate is set to $5e^{-5}$. The number of rounds $T$ is set to 50. Each client trains 1 epoch per round with a batch size of 384. The hyperparameter $\gamma$ in Equation \ref{eq:loss} is set to 0.02 and $\tau$ to 0.05, in Equation \ref{eq:pred}, $\alpha$ is set to 0.01 and $k$ to 5. We use mBERT \cite{devlin2018bert} as the pre-trained language model. We directly take the entity title as the entity description, and following previous work \cite{tong-etal-2022-joint}, translate all Non-English language entities to English. The length of the entity title and the tunable parameters of relation are both set to 12. For clarity, we conduct federated training in more of a ``vertical federated learning'' paradigm, where each round only one client participates in the training, and the clients sequentially upload their local weights one after another following an order consensually agreed. This approach aligns with practical scenarios for multiple KG collaborative training, which most likely would occur among multiple institutes that require a beforehand agreement on the training protocol.

    More elaborate client choosing and weight aggregation designs may further improve performance, which we leave for future work.

    \subsection{Baselines}
    We compare our method with the following baselines: 
    \begin{itemize}
        \item \textbf{Alignment-based methods}: We select two alignment-based methods to show how much aligned entities can do contribution for MKGC. 

        (i) KEnS \cite{chen2020multilingual} encodes all entities and relations for different KGs in a share embedding space, and then completes ensemble inference.
        (ii) SS-AGA \cite{huang2022multilingual} integrates all KGs into a single graph, where alignment is treated as a new type of edge.

        \item \textbf{Alignment-free methods}: These methods do not use aligned entities to transfer knowledge between KGs. They are basically the state-of-the-art KGC methods applied to each KG separately. 
        (i) TransE \cite{bordes2013translating} applies translation-based scoring function for KGC. (ii) RotatE \cite{sun2019rotate} is a rotation-based method. (iii) KG-BERT \cite{yao2019kg} employs a pre-trained language model for text-based KGC. (iv) SimKGC \cite{wang-etal-2022-simkgc} is another text-based KGC model with bi-encoder architecture.    
    \end{itemize}
    We present the results of KenS, SS-AGA, TransE, RotatE, and KG-BERT, as obtained from \cite{huang2022multilingual}. Additionally, we implement SimKGC using mBERT as the foundational language model.
    
    \subsection{Main Results}
    
    The main results in Table \ref{tab:main_results} demonstrate the effectiveness of our proposed FedMKGC framework. The poor performance of alignment-free baselines underscores the necessity of knowledge sharing between multilingual KGs. These methods, deficient in knowledge transfer capabilities, fall short of their alignment-based counterparts. In contrast, FedMKGC employs a language model as a subtle conduit for knowledge, delivering robust results even in the absence of aligned entities. It surpasses all alignment-free baselines by a significant margin. For instance, in low-resource KGs as Greek, it achieves improvements of +17.3\%, +27.2\%, and +24.9\% in Hits@1, Hits@10, and MRR respectively over the best-performing baseline.
    These results distinctly validate the efficacy of implicit knowledge dissemination via the language model.
    
    Moreover, it's noteworthy that FedMKGC not only matches but occasionally outperforms methods relying on explicitly aligned entities. This suggests that FedMKGC is adept at knowledge transfer, rivaling the efficiency of methods dependent on time-consuming entity alignment annotation, and without the accompanying privacy challenges.

    \section{Analysis}
    \begin{figure*}[t]
        \centering
        \includegraphics[width=0.95\textwidth]{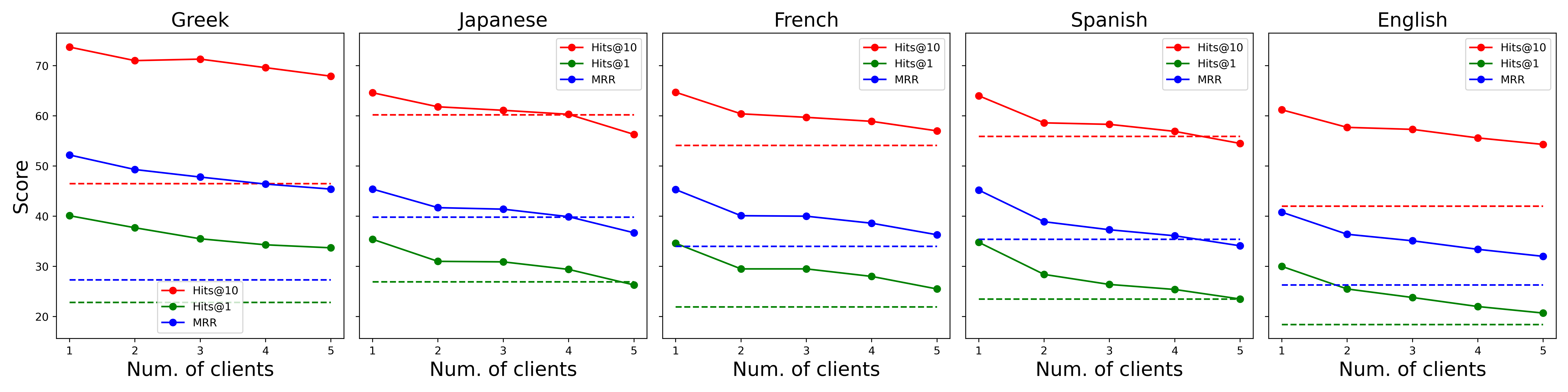} 
        \caption{Influence of the number of clients per round. The x-axis represents the number of clients per round, and the y-axis represents the performance of FedMKGC. The dashed line indicates the score of the best baseline.}
        \label{fig:num_of_clients}
    \end{figure*}
    \begin{table*}[h]
        \caption{Ablation study. All metrics are reported in percentage (\%). TR denotes textual relation. WA denotes weight aggregation. DA denotes data aggregation, OL denotes original language.} 
        \label{tab:ablation}
        \centering
        \setlength\tabcolsep{2.0pt}
        \begin{tabular}{l |  ccc | ccc | ccc | ccc | ccc}
            \toprule
            \multirow{2}*{Model}  & \multicolumn{3}{c}{Greek} & \multicolumn{3}{c}{Japanese} & \multicolumn{3}{c}{French} & \multicolumn{3}{c}{Spanish} &  \multicolumn{3}{c}{English} \\
            \cline{2-4} \cline{5-7} \cline{8-10} \cline{11-13} \cline{14-16}
            & H@1 & H@10 & MRR & H@1 & H@10 & MRR & H@1 & H@10 & MRR & H@1 & H@10 & MRR & H@1 & H@10 & MRR \\
            \midrule 
            \textbf{FedMKGC} & \textbf{40.1}	& \textbf{73.7}	&	\textbf{52.2}	&	\textbf{35.4}	&	\textbf{64.6}	&	\textbf{45.4}	&	\textbf{34.6}	&	\textbf{64.7}	&	\textbf{45.3}	&	\textbf{34.8}	&	\textbf{64.0}	&	\textbf{45.2}	&	\textbf{30.0}	&	\textbf{61.2}	&	\textbf{40.8}  \\
            $\text{FedMKGC}_\text{TR}$  & 38.7	& 73.3 & 	50.7	& 34.4	& 64.3	& 44.8	& 34.0	& 64.4	& 44.7	& 33.2	& 63.6	& 43.9	& 28.8	& 60.4	& 39.7 \\
            $\text{FedMKGC}_\text{WA}$ & 33.7	& 67.9	& 45.4	& 26.3	& 56.3	& 36.7	& 25.5	& 57.0	& 36.3	& 23.5 & 	54.5	& 34.1	& 20.7	& 54.3	& 32.0 \\
            $\text{FedMKGC}_\text{DA}$ & 36.9	& 71.8 & 	49.3	& 30.9	& 62.0	& 41.9	& 31.0	& 61.9	& 41.7	& 29.0	& 60.1	& 39.7	& 25.0	& 57.3	& 36.0 \\
            $\text{FedMKGC}_\text{OL}$ & 18.3	& 47.1 & 	27.4	& 9.5	& 27.2	& 15.4	& 27.2	& 54.4	& 36.5	& 25.7	& 53.0	& 35.1	& 24.2	& 57.4	& 35.5 \\
            \bottomrule
        \end{tabular}
    \end{table*}

    \subsection{Influence of the Number of Clients per Round}

    We explore how the number of clients per round $m$ affects the performance of FedMKGC. Specifically, based on the number of KGs in our dataset, we adjust $m$ from 1 to 5 and keep the total number of weights received by the server consistent. In each round, From all five KGs, we randomly select a predetermined number of KGs to train their respective local models. Subsequently, the server aggregates these models using an average aggregation mechanism, updates the global model, and then dispatches it to the next round's clients.

    As illustrated in Figure \ref{fig:num_of_clients}, FedMKGC's performance diminishes as the number of clients per round escalates. As highlighted by \citeauthor{shi2023towards}, we posit that this decline can be attributed to the inherent data heterogeneity, given that the data distribution among KGs is non-IID. As detailed in Table \ref{tab:dataset}, there is a notable disparity in the KG scale across different clients. Such a straightforward weighted averaging approach might yield suboptimal global models. Consequently, increasing clients per round accentuates the skewed data distribution, adversely impacting FedMKGC's performance. Although devising a more intricate client selection and weight aggregation approach could potentially alleviate the non-IID challenge, this is beyond the scope of our current study and is earmarked for future work.

    The main advantage of increasing the number of clients per round is the ability to parallelize the training process, significantly curtailing training time. For instance, with ample computing resources, concurrently training five clients can expedite the process by nearly fivefold compared to a singular client. Hence, the choice of the number of clients per round can be seen as a trade-off between training time and performance.

    Furthermore, with the number of clients $m < 5$, FedMKGC consistently surpasses all baseline models by a significant margin. As a case in point, with four clients per round, FedMKGC attains a 69.6\% Hits@10 on Greek, marking a 23.1\% improvement over the leading baseline. These results underscore the efficacy of our proposed federated learning framework.

    \subsection{Ablation Study}

    In this section, we present an extensive set of experiments to delve deeper into our analysis. Table \ref{tab:ablation} showcases four variants of FedMKGC, designed to assess the impact of: (1) relation representation, (2) aggregation strategy, and (3) language translation, respectively.

    \subsubsection{Relation Representation} 
    To assess the influence of relation representation, we contrast the performance of FedMKGC with $\text{FedMKGC}_\text{TR}$, which adopts schema text in lieu of our proposed tunable parameters for relation representation. As evidenced in Table \ref{tab:ablation}, FedMKGC consistently surpasses $\text{FedMKGC}_\text{TR}$ across all evaluation metrics on all datasets. This underscores the superiority of our parameterized relation representation, reaffirming its potential to offer heightened flexibility, expressiveness, and subsequently, improved performance in multilingual scenarios.

    \subsubsection{Aggregation Strategy}

    In addition to weight aggregation, we investigate another aggregation strategy termed ``data aggregation.'' Contrary to retaining the knowledge graph at each client for training the local language model and solely aggregating the model's weights, data aggregation necessitates that all clients forward their raw knowledge graph data to a centralized server. Subsequently, this server trains a global language model using the aggregated data. Notably, this approach would influence the in-batch negative sampling process, given that the negative samples will now encompass entities from various KGs rather than exclusively from the respective client's KG. We implement the variant with $m = 5$, which is symbolized as $ \text{FedMKGC}_\text{DA} $ in Table \ref{tab:ablation}. For comparison, $ \text{FedMKGC}_\text{WA} $ represents the model with our proposed weight aggregation strategy, maintaining $ m=5 $.

    It is also should be noted that, with the data aggregation strategy, raw data have to be uploaded to the server, which requires a trusted central server or would raise data leaking concerns. Furthermore, the training process cannot be parallelized since there is only one model, the global language model, trained per round.

    From our observations in Table \ref{tab:ablation}, $\text{FedMKGC}_\text{DA}$ outperforms $\text{FedMKGC}_\text{WA}$. These results suggest that data aggregation is adept at addressing the heterogeneous data challenge, given its interaction with a unified data distribution during training. However, a discernible performance gap remains when compared with the proposed FedMKGC. We hypothesize that this discrepancy primarily arises from the distribution mismatch between training and prediction, a consequence of the influenced in-batch negative sampling process. Alleviating this mismatch is plausible by limiting the source of negative samples, which, in fact, is aligned to FedMKGC with $ m = 1 $.

    \subsubsection{Language Translation}

    We introduce a modification of our model, denoted as $\text{FedMKGC}_\text{OL}$. This variant directly employs the entity titles in their native languages as the description. Referring to Table \ref{tab:ablation}, it is evident that the untranslated $\text{FedMKGC}_\text{OL}$ consistently surpasses all baseline models across multiple datasets. Such results underscore the capability of our proposed framework to adeptly consolidate knowledge, even when presented in diverse languages.

    However, we also observe performance disparities between FedMKGC and $\text{FedMKGC}_\text{OL}$. We ascribe these discrepancies to two primary factors:
    (1) Clients that incorporate knowledge graphs in a myriad of languages intensify the challenge of data heterogeneity. Language models, in such scenarios, grapple with comprehending analogous knowledge when it is conveyed in different languages.
    (2) As delineated in the model card of mBERT \cite{devlin2018bert}, there exists a pronounced imbalance in the pre-training data across languages. This skew leads to certain low-resource languages being ``under-represented''. An empirical manifestation of this can also be observed in our experiments: $\text{FedMKGC}_\text{OL}$ markedly lags behind the SimKGC baseline for Japanese data, given that SimKGC benefits from training on translated datasets.

    While the process of translation might introduce noise, representing knowledge from multilingual KGs in a uniform language can attenuate data heterogeneity. Moreover, it can more effectively harness the strengths of pre-trained language model, resulting in superior performance.

    \section{Conclusion}

    In this work, we introduce FedMKGC, a novel federated learning framework designed for multilingual knowledge graph completion. Unlike traditional methods, FedMKGC eschews the transmission of raw data between KGs. Instead, it facilitates implicit knowledge aggregation without compromising privacy or demanding time-consuming data annotations. In FedMKGC, each KG acts as an individual client which employs a text-based knowledge representation learning method to embed knowledge into its respective local language model. By amalgamating the parameters of these local models, knowledge from various KGs is seamlessly aggregated into a global language model. This approach not only ensures the preservation of data privacy but also obviates the need for potentially risky data sharing and labor-intensive alignment annotations. Compared to existing baselines, FedMKGC demonstrates remarkable performance enhancement. We provide in-depth analysis to elucidate the model's architecture and substantiate our design choices.

\bibliography{custom}

\begin{thebibliography}{33}
\providecommand{\natexlab}[1]{#1}

\bibitem[{Balazevic, Allen, and Hospedales(2019)}]{balazevic-etal-2019-tucker}
Balazevic, I.; Allen, C.; and Hospedales, T.~M. 2019.
\newblock TuckER: Tensor Factorization for Knowledge Graph Completion.
\newblock In Inui, K.; Jiang, J.; Ng, V.; and Wan, X., eds., \emph{Proceedings of the 2019 Conference on Empirical Methods in Natural Language Processing and the 9th International Joint Conference on Natural Language Processing, {EMNLP-IJCNLP} 2019, Hong Kong, China, November 3-7, 2019}, 5184--5193. Association for Computational Linguistics.

\bibitem[{Bordes et~al.(2013)Bordes, Usunier, Garc{\'{\i}}a{-}Dur{\'{a}}n, Weston, and Yakhnenko}]{bordes2013translating}
Bordes, A.; Usunier, N.; Garc{\'{\i}}a{-}Dur{\'{a}}n, A.; Weston, J.; and Yakhnenko, O. 2013.
\newblock Translating Embeddings for Modeling Multi-relational Data.
\newblock In Burges, C. J.~C.; Bottou, L.; Ghahramani, Z.; and Weinberger, K.~Q., eds., \emph{Advances in Neural Information Processing Systems 26: 27th Annual Conference on Neural Information Processing Systems 2013. Proceedings of a meeting held December 5-8, 2013, Lake Tahoe, Nevada, United States}, 2787--2795.

\bibitem[{Chen et~al.(2017)Chen, Tian, Yang, and Zaniolo}]{chen2016multilingual}
Chen, M.; Tian, Y.; Yang, M.; and Zaniolo, C. 2017.
\newblock Multilingual Knowledge Graph Embeddings for Cross-lingual Knowledge Alignment.
\newblock In Sierra, C., ed., \emph{Proceedings of the Twenty-Sixth International Joint Conference on Artificial Intelligence, {IJCAI} 2017, Melbourne, Australia, August 19-25, 2017}, 1511--1517. ijcai.org.

\bibitem[{Chen et~al.(2021)Chen, Zhang, Yuan, Jia, and Chen}]{FedE}
Chen, M.; Zhang, W.; Yuan, Z.; Jia, Y.; and Chen, H. 2021.
\newblock FedE: Embedding Knowledge Graphs in Federated Setting.
\newblock In \emph{IJCKG'21: The 10th International Joint Conference on Knowledge Graphs, Virtual Event, Thailand, December 6 - 8, 2021}, 80--88. {ACM}.

\bibitem[{Chen et~al.(2022)Chen, Zhang, Yuan, Jia, and Chen}]{CHEN2022109459}
Chen, M.; Zhang, W.; Yuan, Z.; Jia, Y.; and Chen, H. 2022.
\newblock Federated knowledge graph completion via embedding-contrastive learning.
\newblock \emph{Knowl. Based Syst.}, 252: 109459.

\bibitem[{Chen et~al.(2020)Chen, Chen, Fan, Uppunda, Sun, and Zaniolo}]{chen2020multilingual}
Chen, X.; Chen, M.; Fan, C.; Uppunda, A.; Sun, Y.; and Zaniolo, C. 2020.
\newblock Multilingual Knowledge Graph Completion via Ensemble Knowledge Transfer.
\newblock In Cohn, T.; He, Y.; and Liu, Y., eds., \emph{Findings of the Association for Computational Linguistics: {EMNLP} 2020, Online Event, 16-20 November 2020}, volume {EMNLP} 2020 of \emph{Findings of {ACL}}, 3227--3238. Association for Computational Linguistics.

\bibitem[{Devlin et~al.(2019)Devlin, Chang, Lee, and Toutanova}]{devlin2018bert}
Devlin, J.; Chang, M.; Lee, K.; and Toutanova, K. 2019.
\newblock {BERT:} Pre-training of Deep Bidirectional Transformers for Language Understanding.
\newblock In Burstein, J.; Doran, C.; and Solorio, T., eds., \emph{Proceedings of the 2019 Conference of the North American Chapter of the Association for Computational Linguistics: Human Language Technologies, {NAACL-HLT} 2019, Minneapolis, MN, USA, June 2-7, 2019, Volume 1 (Long and Short Papers)}, 4171--4186. Association for Computational Linguistics.

\bibitem[{Hu et~al.(2022)Hu, Xu, Yu, Wang, Yang, Zhu, Chang, and Sun}]{hu-etal-2022-empowering}
Hu, Z.; Xu, Y.; Yu, W.; Wang, S.; Yang, Z.; Zhu, C.; Chang, K.-W.; and Sun, Y. 2022.
\newblock Empowering Language Models with Knowledge Graph Reasoning for Open-Domain Question Answering.
\newblock In \emph{Proceedings of the 2022 Conference on Empirical Methods in Natural Language Processing}, 9562--9581. Abu Dhabi, United Arab Emirates: Association for Computational Linguistics.

\bibitem[{Huang et~al.(2022)Huang, Li, Jiang, Cao, Lu, Yin, Subbian, Sun, and Wang}]{huang2022multilingual}
Huang, Z.; Li, Z.; Jiang, H.; Cao, T.; Lu, H.; Yin, B.; Subbian, K.; Sun, Y.; and Wang, W. 2022.
\newblock Multilingual Knowledge Graph Completion with Self-Supervised Adaptive Graph Alignment.
\newblock In Muresan, S.; Nakov, P.; and Villavicencio, A., eds., \emph{Proceedings of the 60th Annual Meeting of the Association for Computational Linguistics (Volume 1: Long Papers), {ACL} 2022, Dublin, Ireland, May 22-27, 2022}, 474--485. Association for Computational Linguistics.

\bibitem[{Kingma and Ba(2015)}]{DBLP:journals/corr/KingmaB14}
Kingma, D.~P.; and Ba, J. 2015.
\newblock Adam: {A} Method for Stochastic Optimization.
\newblock In Bengio, Y.; and LeCun, Y., eds., \emph{3rd International Conference on Learning Representations, {ICLR} 2015, San Diego, CA, USA, May 7-9, 2015, Conference Track Proceedings}.

\bibitem[{Li and Liang(2021)}]{li-liang-2021-prefix}
Li, X.~L.; and Liang, P. 2021.
\newblock Prefix-Tuning: Optimizing Continuous Prompts for Generation.
\newblock In \emph{Proceedings of the 59th Annual Meeting of the Association for Computational Linguistics and the 11th International Joint Conference on Natural Language Processing (Volume 1: Long Papers)}, 4582--4597. Online: Association for Computational Linguistics.

\bibitem[{Lin et~al.(2015)Lin, Liu, Sun, Liu, and Zhu}]{lin2015learning}
Lin, Y.; Liu, Z.; Sun, M.; Liu, Y.; and Zhu, X. 2015.
\newblock Learning Entity and Relation Embeddings for Knowledge Graph Completion.
\newblock In Bonet, B.; and Koenig, S., eds., \emph{Proceedings of the Twenty-Ninth {AAAI} Conference on Artificial Intelligence, January 25-30, 2015, Austin, Texas, {USA}}, 2181--2187. {AAAI} Press.

\bibitem[{Lv et~al.(2022)Lv, Lin, Cao, Hou, Li, Liu, Li, and Zhou}]{lv2022pre}
Lv, X.; Lin, Y.; Cao, Y.; Hou, L.; Li, J.; Liu, Z.; Li, P.; and Zhou, J. 2022.
\newblock Do Pre-trained Models Benefit Knowledge Graph Completion? {A} Reliable Evaluation and a Reasonable Approach.
\newblock In Muresan, S.; Nakov, P.; and Villavicencio, A., eds., \emph{Findings of the Association for Computational Linguistics: {ACL} 2022, Dublin, Ireland, May 22-27, 2022}, 3570--3581. Association for Computational Linguistics.

\bibitem[{McMahan et~al.(2017)McMahan, Moore, Ramage, Hampson, and y~Arcas}]{mcmahan2017communication}
McMahan, B.; Moore, E.; Ramage, D.; Hampson, S.; and y~Arcas, B.~A. 2017.
\newblock Communication-Efficient Learning of Deep Networks from Decentralized Data.
\newblock In Singh, A.; and Zhu, X.~J., eds., \emph{Proceedings of the 20th International Conference on Artificial Intelligence and Statistics, {AISTATS} 2017, 20-22 April 2017, Fort Lauderdale, FL, {USA}}, volume~54 of \emph{Proceedings of Machine Learning Research}, 1273--1282. {PMLR}.

\bibitem[{Nickel, Rosasco, and Poggio(2016)}]{10.5555/3016100.3016172}
Nickel, M.; Rosasco, L.; and Poggio, T. 2016.
\newblock Holographic Embeddings of Knowledge Graphs.
\newblock In \emph{Proceedings of the Thirtieth AAAI Conference on Artificial Intelligence}, AAAI'16, 1955–1961. AAAI Press.

\bibitem[{Nickel, Tresp, and Kriegel(2011)}]{nickel2011three}
Nickel, M.; Tresp, V.; and Kriegel, H. 2011.
\newblock A Three-Way Model for Collective Learning on Multi-Relational Data.
\newblock In Getoor, L.; and Scheffer, T., eds., \emph{Proceedings of the 28th International Conference on Machine Learning, {ICML} 2011, Bellevue, Washington, USA, June 28 - July 2, 2011}, 809--816. Omnipress.

\bibitem[{Peng et~al.(2021)Peng, Li, Song, Zheng, and Li}]{Peng-2021-DPFKGE}
Peng, H.; Li, H.; Song, Y.; Zheng, V.~W.; and Li, J. 2021.
\newblock Differentially Private Federated Knowledge Graphs Embedding.
\newblock In Demartini, G.; Zuccon, G.; Culpepper, J.~S.; Huang, Z.; and Tong, H., eds., \emph{{CIKM} '21: The 30th {ACM} International Conference on Information and Knowledge Management, Virtual Event, Queensland, Australia, November 1 - 5, 2021}, 1416--1425. {ACM}.

\bibitem[{Shi et~al.(2023)Shi, Liang, Zhang, Tan, and Bai}]{shi2023towards}
Shi, Y.; Liang, J.; Zhang, W.; Tan, V.; and Bai, S. 2023.
\newblock Towards Understanding and Mitigating Dimensional Collapse in Heterogeneous Federated Learning.
\newblock In \emph{The Eleventh International Conference on Learning Representations}.

\bibitem[{Singh et~al.(2021)Singh, Chakrabarti, Jain, Choudhury, and Mausam}]{singh2021multilingual}
Singh, H.; Chakrabarti, S.; Jain, P.; Choudhury, S.~R.; and Mausam. 2021.
\newblock Multilingual Knowledge Graph Completion With Joint Relation and Entity Alignment.
\newblock In Chen, D.; Berant, J.; McCallum, A.; and Singh, S., eds., \emph{3rd Conference on Automated Knowledge Base Construction, {AKBC} 2021, Virtual, October 4-8, 2021}.

\bibitem[{Sun et~al.(2021)Sun, Wang, Feng, Ding, Pang, Shang, Liu, Chen, Zhao, Lu, Liu, Wu, Gong, Liang, Shang, Sun, Liu, Ouyang, Yu, Tian, Wu, and Wang}]{sun2021ernie}
Sun, Y.; Wang, S.; Feng, S.; Ding, S.; Pang, C.; Shang, J.; Liu, J.; Chen, X.; Zhao, Y.; Lu, Y.; Liu, W.; Wu, Z.; Gong, W.; Liang, J.; Shang, Z.; Sun, P.; Liu, W.; Ouyang, X.; Yu, D.; Tian, H.; Wu, H.; and Wang, H. 2021.
\newblock ERNIE 3.0: Large-scale Knowledge Enhanced Pre-training for Language Understanding and Generation.
\newblock arXiv:2107.02137.

\bibitem[{Sun et~al.(2019)Sun, Deng, Nie, and Tang}]{sun2019rotate}
Sun, Z.; Deng, Z.; Nie, J.; and Tang, J. 2019.
\newblock RotatE: Knowledge Graph Embedding by Relational Rotation in Complex Space.
\newblock In \emph{7th International Conference on Learning Representations, {ICLR} 2019, New Orleans, LA, USA, May 6-9, 2019}. OpenReview.net.

\bibitem[{Tan et~al.(2023)Tan, Chen, Feng, Zhang, Zheng, Li, and Luo}]{tan2023kracl}
Tan, Z.; Chen, Z.; Feng, S.; Zhang, Q.; Zheng, Q.; Li, J.; and Luo, M. 2023.
\newblock {KRACL:} Contrastive Learning with Graph Context Modeling for Sparse Knowledge Graph Completion.
\newblock In Ding, Y.; Tang, J.; Sequeda, J.~F.; Aroyo, L.; Castillo, C.; and Houben, G., eds., \emph{Proceedings of the {ACM} Web Conference 2023, {WWW} 2023, Austin, TX, USA, 30 April 2023 - 4 May 2023}, 2548--2559. {ACM}.

\bibitem[{Tong et~al.(2022)Tong, Nguyen, Huynh, Nguyen, Nguyen, and Niepert}]{tong-etal-2022-joint}
Tong, V.; Nguyen, D.~Q.; Huynh, T.~T.; Nguyen, T.~T.; Nguyen, Q. V.~H.; and Niepert, M. 2022.
\newblock Joint Multilingual Knowledge Graph Completion and Alignment.
\newblock In Goldberg, Y.; Kozareva, Z.; and Zhang, Y., eds., \emph{Findings of the Association for Computational Linguistics: EMNLP 2022}, 4646--4658. Abu Dhabi, United Arab Emirates: Association for Computational Linguistics.

\bibitem[{Wang et~al.(2021)Wang, Shen, Long, Zhou, Wang, and Chang}]{wang2021structure}
Wang, B.; Shen, T.; Long, G.; Zhou, T.; Wang, Y.; and Chang, Y. 2021.
\newblock Structure-Augmented Text Representation Learning for Efficient Knowledge Graph Completion.
\newblock In Leskovec, J.; Grobelnik, M.; Najork, M.; Tang, J.; and Zia, L., eds., \emph{{WWW} '21: The Web Conference 2021, Virtual Event / Ljubljana, Slovenia, April 19-23, 2021}, 1737--1748. {ACM} / {IW3C2}.

\bibitem[{Wang et~al.(2022)Wang, Zhao, Wei, and Liu}]{wang-etal-2022-simkgc}
Wang, L.; Zhao, W.; Wei, Z.; and Liu, J. 2022.
\newblock SimKGC: Simple Contrastive Knowledge Graph Completion with Pre-trained Language Models.
\newblock In Muresan, S.; Nakov, P.; and Villavicencio, A., eds., \emph{Proceedings of the 60th Annual Meeting of the Association for Computational Linguistics (Volume 1: Long Papers), {ACL} 2022, Dublin, Ireland, May 22-27, 2022}, 4281--4294. Association for Computational Linguistics.

\bibitem[{Wang et~al.(2019)Wang, He, Cao, Liu, and Chua}]{10.1145/3292500.3330989}
Wang, X.; He, X.; Cao, Y.; Liu, M.; and Chua, T.-S. 2019.
\newblock KGAT: Knowledge Graph Attention Network for Recommendation.
\newblock In \emph{Proceedings of the 25th ACM SIGKDD International Conference on Knowledge Discovery \& Data Mining}, KDD '19, 950–958. New York, NY, USA: Association for Computing Machinery.
\newblock ISBN 9781450362016.

\bibitem[{Wang et~al.(2014)Wang, Zhang, Feng, and Chen}]{wang2014knowledge}
Wang, Z.; Zhang, J.; Feng, J.; and Chen, Z. 2014.
\newblock Knowledge Graph Embedding by Translating on Hyperplanes.
\newblock In Brodley, C.~E.; and Stone, P., eds., \emph{Proceedings of the Twenty-Eighth {AAAI} Conference on Artificial Intelligence, July 27 -31, 2014, Qu{\'{e}}bec City, Qu{\'{e}}bec, Canada}, 1112--1119. {AAAI} Press.

\bibitem[{Xie et~al.(2022)Xie, Zhang, Li, Deng, Chen, Xiong, Chen, and Chen}]{xie2022discrimination}
Xie, X.; Zhang, N.; Li, Z.; Deng, S.; Chen, H.; Xiong, F.; Chen, M.; and Chen, H. 2022.
\newblock From Discrimination to Generation: Knowledge Graph Completion with Generative Transformer.
\newblock In Laforest, F.; Troncy, R.; Simperl, E.; Agarwal, D.; Gionis, A.; Herman, I.; and M{\'{e}}dini, L., eds., \emph{Companion of The Web Conference 2022, Virtual Event / Lyon, France, April 25 - 29, 2022}, 162--165. {ACM}.

\bibitem[{Yang et~al.(2015)Yang, Yih, He, Gao, and Deng}]{yang2014embedding}
Yang, B.; Yih, W.; He, X.; Gao, J.; and Deng, L. 2015.
\newblock Embedding Entities and Relations for Learning and Inference in Knowledge Bases.
\newblock In Bengio, Y.; and LeCun, Y., eds., \emph{3rd International Conference on Learning Representations, {ICLR} 2015, San Diego, CA, USA, May 7-9, 2015, Conference Track Proceedings}.

\bibitem[{Yao, Mao, and Luo(2019)}]{yao2019kg}
Yao, L.; Mao, C.; and Luo, Y. 2019.
\newblock {KG-BERT:} {BERT} for Knowledge Graph Completion.
\newblock \emph{CoRR}, abs/1909.03193.

\bibitem[{Yasunaga et~al.(2021)Yasunaga, Ren, Bosselut, Liang, and Leskovec}]{yasunaga-etal-2021-qa}
Yasunaga, M.; Ren, H.; Bosselut, A.; Liang, P.; and Leskovec, J. 2021.
\newblock {QA}-{GNN}: Reasoning with Language Models and Knowledge Graphs for Question Answering.
\newblock In \emph{Proceedings of the 2021 Conference of the North American Chapter of the Association for Computational Linguistics: Human Language Technologies}, 535--546. Online: Association for Computational Linguistics.

\bibitem[{Zhang et~al.(2023)Zhang, Serban, Sun, and Guo}]{zhang2023collective}
Zhang, W.; Serban, O.; Sun, J.; and Guo, Y. 2023.
\newblock Collective Knowledge Graph Completion with Mutual Knowledge Distillation.
\newblock \emph{CoRR}, abs/2305.15895.

\bibitem[{Zhu, Li, and Hu(2023)}]{zhu2023heterogeneous}
Zhu, X.; Li, G.; and Hu, W. 2023.
\newblock Heterogeneous Federated Knowledge Graph Embedding Learning and Unlearning.
\newblock In Ding, Y.; Tang, J.; Sequeda, J.~F.; Aroyo, L.; Castillo, C.; and Houben, G., eds., \emph{Proceedings of the {ACM} Web Conference 2023, {WWW} 2023, Austin, TX, USA, 30 April 2023 - 4 May 2023}, 2444--2454. {ACM}.

\end{thebibliography}
\end{document}